\documentclass[runningheads]{llncs}
\usepackage{comment}
\usepackage[accsupp]{axessibility}
\usepackage{cite}
\usepackage{multirow}
\usepackage{bm}
\usepackage{subcaption}
\usepackage{wrapfig}

\usepackage[pagebackref,breaklinks,colorlinks]{hyperref}

\usepackage{utils}

\makeatletter
\renewcommand\subsection{
  \@startsection{subsection}{2}{\z@}%
  {-14\p@ \@plus -4\p@}
  {4\p@ \@plus 4\p@}
  {\normalfont\normalsize\bfseries\boldmath \rightskip=\z@ \@plus 8em\pretolerance=10000}
}
\renewcommand\subsubsection{
  \@startsection{subsubsection}{3}{\z@}%
  {-14\p@ \@plus -4\p@}
  {-0.5em \@plus -0.22em \@minus -0.1em}
  {\normalfont\normalsize\bfseries\boldmath}
}

\makeatother

\begin{document}

\pagestyle{headings}
\mainmatter

\title{HEAD: HEtero-Assists Distillation for Heterogeneous Object Detectors}
\titlerunning{HEAD: HEtero-Assists Distillation for Heterogeneous Object Detectors}

\author{  
  Luting Wang\textsuperscript{\rm1,3}\quad  
  Xiaojie Li\textsuperscript{\rm 2}\quad 
  Yue Liao\textsuperscript{\rm 1,3}\thanks{Corresponding author (liaoyue.ai@gmail.com)}\quad 
  Zeren Jiang\textsuperscript{\rm 4}\\ 
  Jianlong Wu\textsuperscript{\rm 5}\quad 
  Fei Wang\textsuperscript{\rm 2,6}\quad 
  Chen Qian\textsuperscript{\rm 2}\quad 
  Si Liu\textsuperscript{\rm 1,3} 
}
\authorrunning{Wang \etal}
\institute{
  \textsuperscript{\rm 1}Institute of Artificial Intelligence, Beihang University\quad
  \textsuperscript{\rm 2}SenseTime Research\\ 
  \textsuperscript{\rm 3}Hangzhou Innovation Institute, Beihang University\quad 
  \textsuperscript{\rm 4}ETH Zurich\\
  \textsuperscript{\rm 5}Shandong University\quad 
  \textsuperscript{\rm 6}University of Science and Technology of China \\ \url{https://github.com/LutingWang/HEAD}
}

\maketitle

\begin{abstract}
Conventional knowledge distillation~(KD) methods for object detection mainly concentrate on homogeneous teacher-student detectors. However, the design of a lightweight detector for deployment is often significantly different from a high-capacity detector. Thus, we investigate KD among heterogeneous teacher-student pairs for a wide application. We observe that the core difficulty for heterogeneous KD~(hetero-KD) is the significant semantic gap between the backbone features of heterogeneous detectors due to the different optimization manners. Conventional homogeneous KD~(homo-KD) methods suffer from such a gap and are hard to directly obtain satisfactory performance for hetero-KD. In this paper, we propose the HEtero-Assists Distillation (HEAD) framework, leveraging heterogeneous detection heads as assistants to guide the optimization of the student detector to reduce this gap. In HEAD, the assistant is an additional detection head with the architecture homogeneous to the teacher head attached to the student backbone. Thus, a hetero-KD is transformed into a homo-KD, allowing efficient knowledge transfer from the teacher to the student. Moreover, we extend HEAD into a Teacher-Free HEAD (TF-HEAD) framework when a well-trained teacher detector is unavailable. Our method has achieved significant improvement compared to current detection KD methods. For example, on the MS-COCO dataset, TF-HEAD helps R18 RetinaNet achieve $33.9$ mAP ($+2.2$), while HEAD further pushes the limit to $36.2$ mAP ($+4.5$). 
\keywords{Knowledge Distillation, Object Detection, Heterogeneous.}
\end{abstract}

\section{Introduction}

With the development of deep learning, the performance of object detection has achieved tremendous improvement. However, deploying detectors to edge devices often imposes constraints on the number of parameters, computation, and memory.
Therefore, parameters compression and accuracy boosting are core problems for object detection towards practical application, where knowledge distillation (KD) is one of the most popular solutions.
KD aims at training the compact model (student) by transferring knowledge from a high-capacity model (teacher). 
Recently, with the development of KD methods~\cite{kd, fitnets, one, dml, rkd, fsp} in general vision models, KD in object detection has raised increasing attention.

For a clear presentation, we first give a brief definition for the general architecture of modern CNN based detectors~\cite{retinanet,fcos,faster_rcnn}, where input images are represented as feature maps and then different methods are used to decode detection results from the feature maps. 
In this way, a detector can be divided into a backbone including FPN~\cite{fpn} and a detection head. 
We further define detectors with the same head architecture as homogeneous, otherwise heterogeneous. 

Based on this definition, we summarize object detection KD into two schemes based on the architecture of teacher-student pair: homogeneous KD~(homo-KD) or heterogeneous KD~(hetero-KD). 
The homo-KD scheme usually allows the teacher detector to have a stronger backbone than the student detector but requires same head architectures. 
For instance, R18 RetinaNet~\cite{retinanet} can be taught by R50 RetinaNet by homo-KD, but not R50 RepPoints~\cite{reppoints} or R50 Faster R-CNN~\cite{faster_rcnn}. 
Significant progress has been made in the homo-KD~\cite{fitnets,detection_kd,fgfi,gid, defeat, fkd, fgd}. However, the practical application of homo-KD is limited because the student for deployment and the most powerful teacher is usually designed from different motivations and produce very different heads. 
Therefore, we aim to explore hetero-KD, an essential and significant topic for object detection.

\begin{figure}[t]
  \centering
  \begin{subfigure}[b]{0.24\linewidth}
    \includegraphics[width=\linewidth]{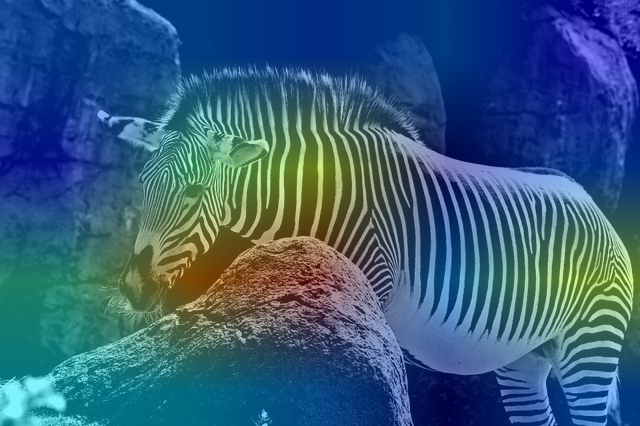}
    \caption{RetinaNet}
  \end{subfigure}
  \hfill
  \begin{subfigure}[b]{0.24\linewidth}
    \includegraphics[width=\linewidth]{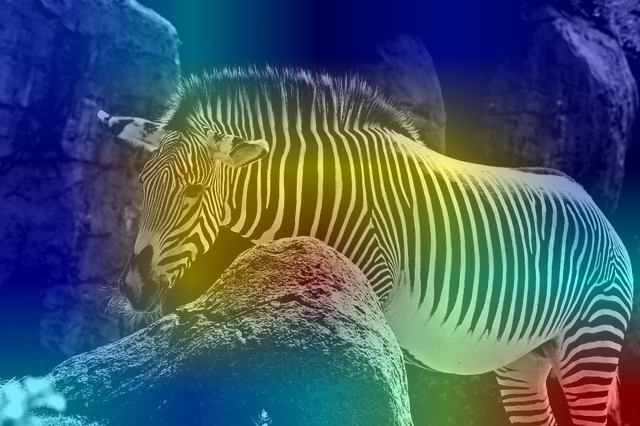}
    \caption{FCOS}
  \end{subfigure}
  \hfill
  \begin{subfigure}[b]{0.24\linewidth}
    \includegraphics[width=\linewidth]{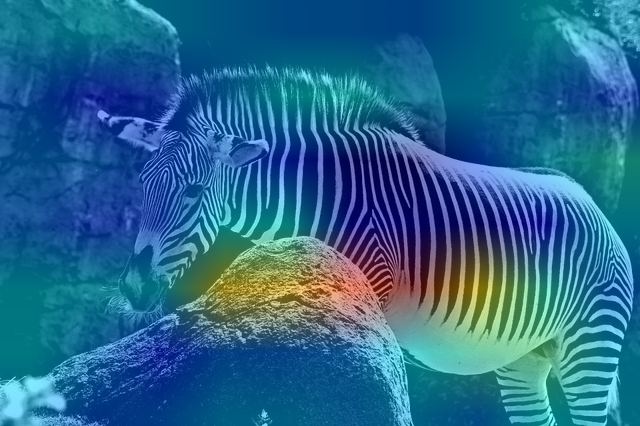}
    \caption{Faster R-CNN}
  \end{subfigure}
  \hfill
  \begin{subfigure}[b]{0.24\linewidth}
    \includegraphics[width=\linewidth]{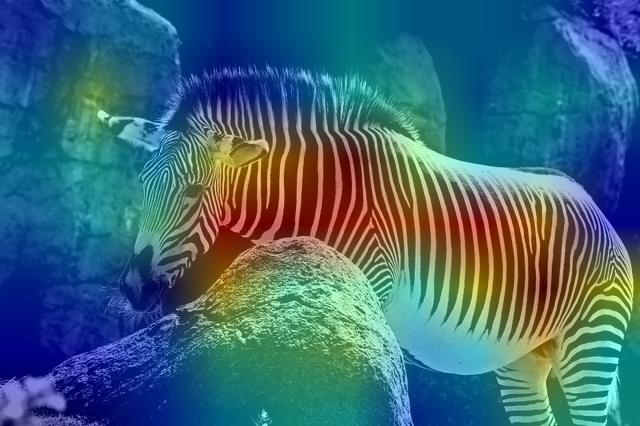}
    \caption{TF-HEAD}
  \end{subfigure}
  \caption{
    Comparison of the activation patterns from different detectors with the same backbone architecture. 
    The intensity of the feature response increases from blue to red. 
    These detectors produce different backbone feature representations. 
    We use RetinaNet as the student and apply TF-HEAD to take advantage of the feature extraction abilities from an FCOS assistant and an R-CNN assistant. 
    As a result, the activation map of TF-HEAD highlights more area of the zebra with higher intensity, indicating that the feature map contains the most information.
  }
  \label{fig:fig1}
\end{figure}

We first present an analysis for heterogeneous detectors. 
We observe that two heterogeneous detectors share the same backbone architecture, while their backbone features representation are still very distinct. As shown in \cref{fig:fig1}, activation maps from different detectors with R50 backbone, \eg Faster R-CNN~\cite{faster_rcnn}, RetinaNet~\cite{retinanet}, and FCOS~\cite{fcos}, are different.
Thus, we argue that the heterogeneous detection heads guide the backbone for different knowledge.
We consider it a significant step for hetero-KD that the student mimics the teacher backbone knowledge.
The intuitive idea is to perform homo-KD methods for heterogeneous detectors directly, but the accuracy improvement of the student is significantly limited. It is mainly because the backbone knowledge discrepancy enlarges the semantic gaps between the teacher and student layers. 

To this end, we design a simple yet effective hetero-KD mechanism, namely HEtero-Assists Distillation (HEAD), to bridge the semantic gap between heterogeneous detectors via an adaptive assistant, thus simplifying to a homo-KD problem. For a specific teacher-student pair, we first design an additional assistant head with identical architecture with the teacher head and attach it to the student backbone.
In this way, we construct a homogeneous detectors pair, \ie the teacher and the student backbone equipped with the assistant.
During training, the assistant and original student head process the student backbone features in parallel. Then, we propose two KD mechanisms, \ie Assistant-based KD~(AKD) and Cross-architecture KD~(CKD), to supervise the assistant and student head learning, respectively.
In AKD, we directly apply the homo-KD to the assistant and the teacher heads since they are homogeneous. Therefore, the high-level knowledge~\cite{fsp}, \ie the information flow for detection, is efficiently transferred from the teacher head to the assistant. Moreover, the teacher backbone knowledge is also transferred to the student backbone through gradient back-propagation from the assistant.
Intuitively, the assistant teaches the student backbone to learn the critical knowledge reproducing the information flow~\cite{fsp} of the teacher. Thus, the semantic gap between heterogeneous detectors is bridged by the assistant. In CKD, we conduct a feature mimic from the student head to the teacher head. CKD plays an auxiliary role to integrate heterogeneous knowledge in the head level to compensate for AKD.  


In practice, it is not always easy to obtain a suitable teacher for a specific student, limiting the application of traditional teacher-based KD methods~\cite{tfkd, labelenc}. Therefore, we further explore a Teacher-Free~\cite{tfkd, Ji2021, Kim2021, Zhang2019} method which accommodates our HEAD to these situations, namely TF-HEAD. 
TF-HEAD works by injecting diverse knowledge into the student, which helps the student to make more accurate predictions without extra computation cost at inference time. Specifically, we use the assistant to process the student backbone features, wihch is the same as HEAD. Since the assistant and the student heads are heterogeneous, they optimize the student backbone differently, thus enriching the knowledge inside. 
Although we train the assistant with ground truth labels, instead of supervision from the teacher head, the performance improvement brought by the assistant module is still significant. To push the limit further, TF-HEAD uses multiple assistants that are heterogeneous to each other.



Extensive experiments demonstrate the effectiveness of our framework. On MS-COCO dataset~\cite{coco}, our HEAD and TF-HEAD methods achieve state-of-the-art performance among teacher-based detection KD methods and teacher-free detection KD methods respectively. Using R50 Faster R-CNN as teacher, the mAP of R18 RetinaNet is elevated from $31.7$ to $36.2$ ($+4.5$) and R18 FCOS from $32.5$ to $36.0$ ($+3.5$). 
Without pretrained teachers, TF-HEAD improves R18 RetinaNet from $31.7$ to $33.9$ ($+2.2$), which demonstrates that simply integrating heterogeneous knowledge helps to train better detectors, making hetero-KD advantageous compared to homo-KD.


  
  

\section{Related Work}

\vspace{-1mm}\subsection{Object Detection}\vspace{-1mm}

Object detection has three paradigms: two-stage~\cite{fast_rcnn, faster_rcnn, mask_rcnn, spp, fpn, rfcn,centernet2}, anchor-based one-stage~\cite{retinanet, ssd, yolo, yolof}, and anchor-free one stage~\cite{fcos, centernet_objects, centernet_keypoint, centernet2,Dong_2020_CVPR}. Two stage detectors use an Region Proposal Network (RPN) to generate Regions of Interest (RoIs) and then adopt a region-wise prediction network (R-CNN head) to predict objects. Although two-stage architectures obtain high accuracy, their complicated pipeline hinders deployment on edge devices. In contrast, one-stage detectors get the classification and the bounding box of targets based on features extracted by the backbone directly, achieving real-time inference. Anchor-based one-stage detectors use dense anchor boxes as proposals to detect targets. However, the number of anchor boxes is far more than targets, which brings much extra computation. Anchor-free detectors learn to predict keypoints and then generate bounding boxes to detect objects without the need for predefined anchors, reaching better performance with less cost. 
The features extracted by different detectors are optimized by different detection heads, which will result in large semantic gaps. Thus, It's hard to mitigate the difference by traditional homo-KD methods. In this work, we introduce the adaptive assistants to effectively bridge the gap between heterogeneous teacher-student pairs.

\vspace{-1mm}\subsection{Knowledge Distillation}\vspace{-1mm}

\noindent\textbf{General KD.}
KD is a technology that helps training compact student models under the supervision of powerful teacher models. 
Hinton \etal~\cite{kd} propose this concept and achieve great performance improvement by training the student with class distributions generated by the teacher.
Extending Hinton's work, more works~\cite{fitnets, atkd, nst, rkd, spkd, cckd, vid, pkt, abkd, ftkd, fsp, pkth, hsakd, lkd} use intermediate representations of the teacher as hints to train the student. 
TAKD~\cite{takd} employs intermediate-sized networks as assistants to improve the effectiveness of KD when the teacher-student capacity gap is large.
Different from TAKD, our approach adopts the assistants to solve the heterogeneity between detector pairs. 

\noindent\textbf{Homogeneous detection KD.}
Chen \etal~\cite{detection_kd} first apply KD to object detection by implementing feature-based and response-based loss for Faster R-CNN. Li \etal~\cite{mimic} apply L2 loss on features sampled by proposals of the student. Wang \etal~\cite{fgfi} find that mimicking features from foreground regions is more important than background and only distills the feature near object anchor locations. DeFeat~\cite{defeat} 
shows that the information of background features is also essential, so foreground and background regions are distilled simultaneously with different factors. LabelEnc~\cite{labelenc} first trains an autoencoder to model the location-category information and then use the label representations to supervise the training of the detectors. 
Although these methods have achieved great success on heterogeneous backbones, heterogeneity between detection architectures are rarely explored due to the large structural and feature semantics differences.

\noindent\textbf{Heterogeneous Detection KD.}
To transfer knowledge between heterogeneous detectors, MimicDet~\cite{mimicdet} introduces a refinement module to an one-stage detection head to imitate the workflow of two-stage heads, and then conducts KD between the aligned features from the teacher and student heads. Although MimicDet improves accuracy, it is hard to transfer the structural modification on the student head to other heterogeneous detectors. Different from MimicDet, our method is more intuitive and flexible. With minor modifications, a variety of heterogeneous detectors can be supported by our framework. 

G-DetKD~\cite{gdetkd} first proposes a general distillation framework for object detectors, which performs soft matching across all pyramid levels to provide the optimal guidance to the student. However, using learned similarity scores to combine features of students at different levels before feature mimicking does not essentially reduce the semantic gap. 
In our work, we attach an assistant, which is same as the teacher detection head, to the student network to learn directly from the teacher. Since the assistant and the teacher have homogeneous detection heads, their feature semantic gap are smaller than the heterogeneous heads, contributing to more efficient knowledge transfer.


\section{Method}

In \cref{subsec:reviewkd}, we briefly review the pipeline of detection KD. 
Then we elaborate our proposed hetero-KD mechanism, HEtero-Assists Distillation (HEAD) in \cref{subsec:head}. 
Finally, we introduce a teacher-free extension of HEAD~(TF-HEAD) in \cref{subsec:tfhead}.

\subsection{Review of Detection KD}
\label{subsec:reviewkd}

We focus on distillation using intermediate features~\cite{fitnets}. 
The distillation loss of the features can be generally formulated as:
\begin{equation}
\label{eq:fitnet}
  \mathcal{L} = \mathcal{D}\left(\mathbf{F}^T, \phi (\mathbf{F}^S)\right),
\end{equation}
where $\mathcal{D}(\cdot)$ is a distillation loss measuring the knowledge difference between the teacher and the student. 
$\phi$ is an adaptation layer to match the dimension of the student’s feature with the teacher. 
$\mathbf{F}^T$ and $\mathbf{F}^S$ are intermediate features of the teacher and the student respectively. 

In this paper, we define $\mathcal{D}(\cdot)$ as the MSE loss. 
The form of $\phi$ depends on the shape of $\mathbf{F}^S$. For three-dimensional features with different number of channels, $1 \times 1$ convolution layers are used. For two-dimensional features with different number of dimensions, $\phi$ represents a linear layer.

\subsection{HEAD}
\label{subsec:head}

\begin{figure}[t]
  \centering
  \includegraphics[width=\linewidth]{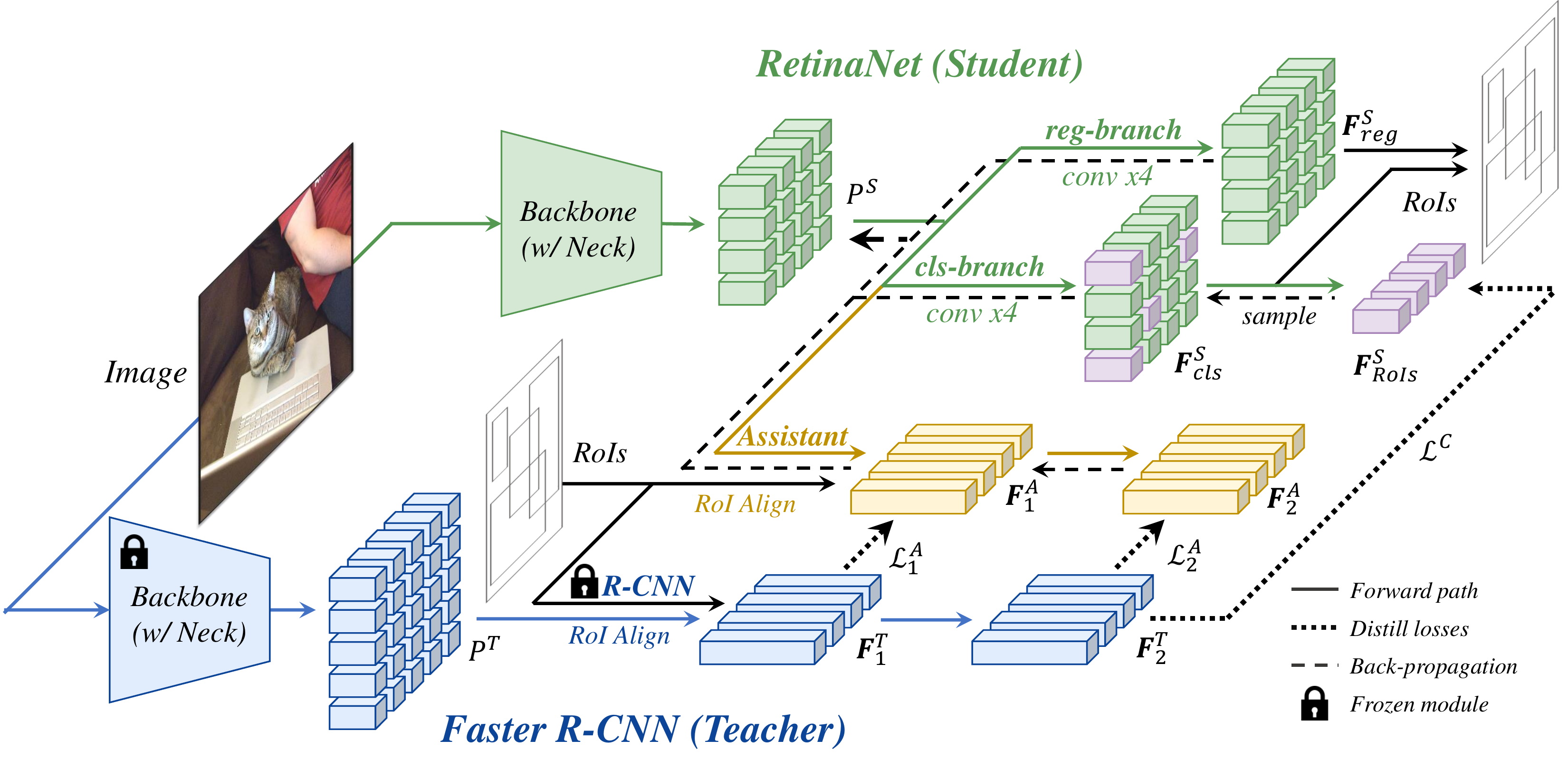}
  \caption{
    Overview of HEAD, where the teacher is Faster R-CNN~\cite{faster_rcnn} and student is RetinaNet~\cite{retinanet}. 
    We construct an assistant homogeneous to the teacher's R-CNN head. 
    We first extract the backbone features of the teacher and the student. 
    The student backbone features are then fed into the original student head and the assistant in parallel. 
    The teacher head processes the teacher backbone features. 
    KD mechanism comprises an AKD between the teacher head and the assistant, and a CKD, where the teacher head directly supervises the student head.
  }
  \label{fig:framework}
\end{figure}

We elaborate the pipeline of HEAD by instantiating an example, where Faster R-CNN~\cite{fast_rcnn} is adopted as the teacher and RetinaNet~\cite{retinanet} is the student.
\Cref{fig:framework} shows the corresponding framework. 
To bridge the semantic gap between teacher and student, HEAD constructs an assistant that is homogeneous with the teacher's R-CNN head. 
The assistant is initialized with the pretrained weight of the teacher's R-CNN head and is trained online with the student. 
Note that HEAD acts on the training phase only, and the assistant is unused during inference. 

Given an image, we first extract the backbone features of student and teacher, denoted as $\mathbf{P}^S \in \mathbb{R}^{C \times H \times W}$ and $\mathbf{P}^T \in \mathbb{R}^{C \times H' \times W'}$ respectively. 
We then follow the original detector pipeline to employ the student head, a RetinaNet~\cite{retinanet} head, to calculate the student loss, denoted as $\mathcal{L}^S_{gt}$. As shown in~\cref{fig:framework}, the student head comprises a regression branch and a classification branch. The student loss is calculated by summarizing losses from both branches:
\begin{equation}
  \mathcal{L}^S_{gt} = \mathcal{L}^S_{reg} + \mathcal{L}^S_{cls}.
\end{equation}
The $\mathcal{L}^S_{gt}$ is calculated following the original RetinaNet. For convinience, we adopt $\mathbf{F}^S_{reg} \in \mathbb{R}^{C \times H \times W}$ and $\mathbf{F}^S_{cls} \in \mathbb{R}^{C \times H \times W}$ to denote the last intermediate feature of the regression branch and the classification branch respectively.

We next introduce the KD mechanism composed of two KD processes, \ie \textit{Assistant-based KD~(AKD)} and \textit{Cross-architecture KD~(CKD)}. AKD is the core of HEAD. We utilize the teacher's R-CNN head and the assistant to process the corresponding backbone features $\mathbf{P}^T$ and $\mathbf{P}^S$, respectively. Meanwhile, we adopt the AKD loss $\mathcal{L}^A$ for the assistant to mimic the intermediate features of the teacher head. For completeness, we employ CKD, where the teacher head directly provides supervision for the student head. The CKD loss is denoted as $\mathcal{L}^C$. During the training phase, the overall loss is
\begin{equation}
  \mathcal{L}^{HEAD} = \mathcal{L}^S_{gt} + \mathcal{L}^A + \mathcal{L}^C.
\end{equation}
Our HEAD framework is not restricted to distillation between one-stage and two-stage detectors but can be applied to a wide range of heterogeneous detectors.

\vspace{1mm}\noindent\textbf{Assistant-based KD.}
When the teacher is two-stage, the teacher head and the assistant perform RoI Align~\cite{mask_rcnn} on $\mathbf{P}^T$ and $\mathbf{P}^S$ respectively with a precomputed set of RoIs. 
For a two-stage student, the output of the student's RPN is used as the precomputed RoIs. 
For one-stage students without RPN, we take the output of the student head as a substitution. 
For the example in \cref{fig:framework}, we convert the classification logits of each anchor to a class-agnostic objectness logit and follow the original RPN protocol to generate RoIs. 
Additionally, we denote the number of RoIs as $N$.

We feed the backbone features $\mathbf{P}^T$ and $\mathbf{P}^S$ (or the RoI Aligned features) into the teacher head and the assistant, respectively. 
The intermediate features of the teacher head and the assistant is respectively denoted as $\mathbf{F}^T_1, \mathbf{F}^T_2, \ldots, \mathbf{F}^T_L$ and $\mathbf{F}^A_1, \mathbf{F}^A_2, \ldots, \mathbf{F}^A_L$. 
$L$ indicates the number of intermediate features. 
In \cref{fig:framework}, the R-CNN head is composed of two linear layers.
We use the outputs of both layers for KD, thus setting $L$ to $2$. 
Finally, since the teacher head is homogeneous with the assistant, we simply apply KD between intermediate features pairs
\begin{equation}
  \mathcal{L}^A_l = \mathcal{D}\left(\mathbf{F}^T_l, \phi \left(\mathbf{F}^S_l\right)\right).
\end{equation}
For simplicity, we use MSE loss as $\mathcal{D}(\cdot)$ and a linear layer as $\phi$. 

Besides the supervision from the teacher, we also use ground truth labels to supervise the assistant. 
In \cref{fig:framework}, we follow the original Faster R-CNN to use the standard Cross-Entropy loss and L1 loss to supervise the classification and regression output of the assistant. 
In general, the ground truth loss for the assistant $\mathcal{L}^A_{gt}$ is same as the teacher, so that the assistant learns from the ground truth labels when the teacher makes mistakes. 
The total AKD loss is defined as
\begin{equation}
    \mathcal{L}^A = \mathcal{L}^A_{gt} + \frac{\lambda^A }{L}\sum^L_{l = 1} \mathcal{L}^A_l,
\end{equation}
where $\lambda^A$ represents the loss weight.

Intuitively, $\mathcal{L}^A$ requires the assistant to reproduce the reasoning process of the teacher. 
For this goal, the student backbone needs to capture enough information in $\mathbf{P}^S$. 
Therefore, the assistant optimizes the student backbone via gradient back-propagation, so that the student backbone learns the knowledge that is critical for the assistant to reproduce the teacher's reasoning process. 

\vspace{1mm}\noindent\textbf{Cross-architecture KD.}
Though the AKD is effective and universal, it only distills knowledge into the student backbone. 
Hence, we design the CKD to further improve the student performance via direct supervisions from the teacher head to the student head. 
As shown in \cref{fig:framework}, we apply the CKD loss $\mathcal{L}^C$ between the teacher's R-CNN head and the student's RetinaNet head. 
Firstly, the teacher's R-CNN head generates a set of sparse RoI features $\mathbf{F}^T_1 \in \mathbb{R}^{N \times C'}$, while the student's RetinaNet head generates a series of dense anchor features $\mathbf{F}^S_{cls} \in \mathbb{R}^{C \times H \times W}$. 
$C'$ represents the dimensions of the R-CNN head's hidden layer. 
We ignore the regression feature $\mathbf{F}^S_{reg}$ following G-DetKD~\cite{gdetkd}. 
Secondly, inspired by MimicDet~\cite{mimicdet}, we trace back to the original anchors of each RoI. 
Thirdly, since each anchor corresponds to a pixel on $\mathbf{F}^S_{cls}$, we sample these pixel features to form $\mathbf{F}^S_{RoIs} \in \mathbb{R}^{N \times C}$. 
Thereafter, we use \cref{eq:fitnet} to perform CKD
\begin{equation}
  \mathcal{L}^C = \lambda^C \mathcal{D}\left(\mathbf{F}^T_1, \phi (\mathbf{F}^S_{RoIs})\right),
\end{equation}
where $\lambda^C$ is the loss weight, $\mathcal{D}(\cdot)$ is MSE loss, and $\phi$ is a linear layer mapping from $C$-dimensional features to $C'$-dimensional features.

If the teacher and the student are both one-stage or both two-stage, CKD simply applies MSE loss between the corresponding intermediate features of the teacher head and the student head. 
Suppose the teacher is FCOS~\cite{fcos} and the student is RetinaNet~\cite{retinanet}. 
Let $\mathbf{F}^T_{FCOS}$ denote the last intermediate feature of the classification branch of the FCOS head. 
Then the CKD loss is denoted as
\begin{equation}
  \mathcal{L}^C_{FCOS} = \lambda^C_{FCOS} \mathcal{D}\left(\mathbf{F}^T_{FCOS}, \phi\left(\mathbf{F}^S_{cls}\right)\right),
\end{equation}
where $\lambda^C_{FCOS}$ is the loss weight, $\mathcal{D}(\cdot)$ is MSE loss and $\phi$ is a $1 \times 1$ convolution layer. 
For both teacher and student with a two-stage pipeline, the only difference is that the adaptation layer $\phi$ uses a linear layer.

\subsection{TF-HEAD}
\label{subsec:tfhead}

\begin{figure}[t]
  \centering
  \includegraphics[width=0.9\linewidth]{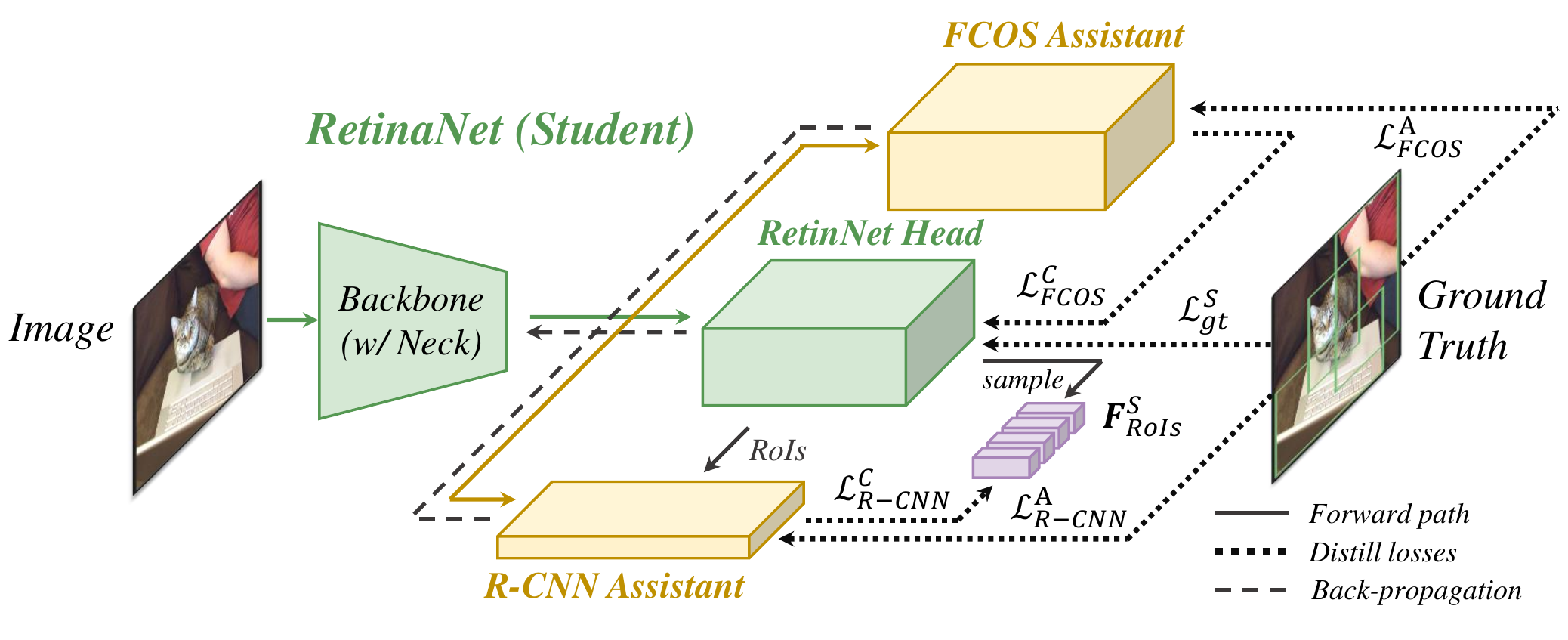}
  \caption{An example of the TF-HEAD training pipeline. We use R-CNN head~\cite{faster_rcnn} and FCOS head~\cite{fcos} as two assistants to guide the RetinaNet~\cite{retinanet}.}
  \label{fig:tf-framework}\vspace{-3mm}
\end{figure}

In practice, suitable teachers for a specific student are not always available.
Traditional teacher-based KD methods, including our HEAD, fail in such situations. 
Therefore, we extend our HEAD to a teacher-free method, namely TF-HEAD.
TF-HEAD is designed based on an experimental observation. 
As shown in \cref{fig:fig1}, different detectors~\cite{faster_rcnn, retinanet, fcos} have distinct activation maps, demonstrating diverse knowledge in the backbones.
Heterogeneous detection architectures incorporate different human priors and adopt various backbone optimization manners.
Therefore, the heterogeneous detectors produce different knowledge. 

Even without the pretrained teacher in HEAD, we observe that the assistant can still learn from the ground truth labels.
Based on this, we devise a teacher-free KD mechanism, TF-HEAD. 
As shown in \cref{fig:tf-framework}, TF-HEAD uses assistants to transfer knowledge from the heterogeneous detectors to the student. 
Note that we allow more than one assistant in TF-HEAD.
While the TF-HEAD framework is universal, we use the example in \cref{fig:tf-framework} to show its application, where the FCOS head and the R-CNN head are adopted to teach the RetinaNet head. 

For each assistant, we denote its ground truth loss and CKD loss as $\mathcal{L}^A_{\star}$ and $\mathcal{L}^C_{\star}$, respectively, as described in \cref{subsec:head}. 
$\star$ indicates the name of the assistant. 
For the example in \cref{fig:tf-framework}, the overall loss is
\begin{equation}
  \mathcal{L}^{TF-HEAD} = \mathcal{L}^S_{gt} + \mathcal{L}^A_{FCOS} + \mathcal{L}^C_{FCOS} + \mathcal{L}^A_{R-CNN} + \mathcal{L}^C_{R-CNN}.
\end{equation}
More generally, the overall loss of our TF-HEAD framework is
\begin{equation}
  \label{eq:tf_head_loss}
  \mathcal{L}^{TF-HEAD} = \mathcal{L}^S_{gt} + \sum_{\star} \left(\mathcal{L}^A_{\star} + \mathcal{L}^C_{\star} \right),
\end{equation}
where $\mathcal{L}^S_{gt}$ is the ground truth loss of the student detector. 
We represent the loss weights of assistant $\star$ as $\lambda^A_{\star}$ and $\lambda^C_{\star}$. 

\section{Experiments}

Experiments are conducted on the COCO $2017$ dataset~\cite{coco} using the mean Average Precision (mAP) metric. 
We adopt the default $120$k/$5$k split for training and validation. 
All distillation loss $\mathcal{D}(\cdot)$ takes the form of MSE. 
The adaptation layer $\phi$ is either a $1 \times 1$ convolution layer or a linear layer, depending on the shape of its input. 
For AKD, we set $\lambda^A$ to $5$. 
If the student head and the assistant are both one-stage heads, the CKD loss weight $\lambda^C$ is set to $1$, otherwise $2$.

Training is conducted on $8$ GPUs with batch size $16$ in total.
We use stochastic gradient descent (SGD) optimizer with $0.9$ momentum and $0.0001$ weight decay. 
1x ($12$ epochs) training schedule is used.
At the $8$\textsuperscript{th} and $11$\textsuperscript{th} epochs, the learning rate is divided by $10$
The initial learning rate is $0.01$ for one-stage detectors and $0.02$ for two-stage detectors.
The shorter side of the input image is scaled to $640$-$800$ pixels, the longer side is scaled to $1333$ pixels. 

\subsection{Main Results}

\begin{table}[t]
\caption{Comparison with homogeneous KD methods. \dag~indicates that only Assistant-based KD losses are used in HEAD.}
\label{tab:homo_sota}
\begin{center}
\begin{tabular}{cc|ccc|ccc}
\hline
\multicolumn{2}{c|}{Student Backbone}                                                                                                                                     & \multicolumn{3}{c|}{R18~\cite{resnet}}                                         & \multicolumn{3}{c}{MNV2~\cite{mobilenetv2}}                                    \\ \hline
\multicolumn{1}{c|}{Method}                                     & Teacher                                                                                                 & \multicolumn{1}{c}{mAP}    & mAP\textsubscript{50} & mAP\textsubscript{75} & \multicolumn{1}{c}{mAP}    & mAP\textsubscript{50} & mAP\textsubscript{75} \\ \hline\hline
\multicolumn{1}{c|}{RetinaNet~\cite{retinanet}}                 & -                                                                                                       & \multicolumn{1}{c}{31.7} & 49.5                  & 33.5                  & \multicolumn{1}{c}{28.5} & 44.8                  & 29.9                  \\ \hline
\multicolumn{1}{c|}{FitNet~\cite{fitnets}}                      & \multirow{5}{*}{\begin{tabular}[c]{@{}c@{}}R50\\ Faster R-CNN\\ (40.3)~\cite{faster_rcnn}\end{tabular}} & 34.1                     & 52.2                  & 36.0                  & 31.6                     & 48.7                  & 33.5                  \\
\multicolumn{1}{c|}{FGFI~\cite{fgfi}}                           &                                                                                                         & 34.4                     & 52.2                  & 36.4                  & 30.8                     & 47.3                  & 32.6                  \\
\multicolumn{1}{c|}{DeFeat~\cite{defeat}}                       &                                                                                                         & 34.1                     & 52.1                  & 36.3                  & 31.1                     & 47.9                  & 32.7                  \\
\multicolumn{1}{c|}{FGD~\cite{fgd}}                             &                                                                                                         & 34.4                     & 52.6                  & 36.7                  & 31.9                     & 48.9                  & 34.0                  \\
\multicolumn{1}{c|}{\textbf{HEAD\textsuperscript{\dag} (ours)}} &                                                                                                         & \textbf{35.5}            & \textbf{54.5}         & \textbf{37.9}         & \textbf{32.5}            & \textbf{50.4}         & \textbf{34.4}         \\ \hline
\multicolumn{1}{c|}{FitNet~\cite{fitnets}}                      & \multirow{5}{*}{\begin{tabular}[c]{@{}c@{}}R50\\ RepPoints\\ (38.6)~\cite{reppoints}\end{tabular}}      & 31.5                     & 49.0                  & 33.3                  & 27.5                     & 43.2                  & 28.9                  \\
\multicolumn{1}{c|}{FGFI~\cite{fgfi}}                           &                                                                                                         & 33.1                     & 50.8                  & 35.3                  & 29.1                     & 45.1                  & 30.8                  \\
\multicolumn{1}{c|}{DeFeat~\cite{defeat}}                       &                                                                                                         & 30.9                     & 48.0                  & 32.8                  & 28.2                     & 44.2                  & 29.6                  \\
\multicolumn{1}{c|}{FGD~\cite{fgd}}                             &                                                                                                         & 31.3                     & 48.6                  & 33.2                  & 28.3                     & 44.3                  & 30.0                  \\
\multicolumn{1}{c|}{\textbf{HEAD\textsuperscript{\dag} (ours)}} &                                                                                                         & \textbf{34.2}            & \textbf{52.4}         & \textbf{36.6}         & \textbf{30.5}            & \textbf{47.1}         & \textbf{32.3}         \\ \hline\hline
\multicolumn{1}{c|}{FCOS~\cite{fcos}}                           & -                                                                                                       & 32.5                     & 50.9                  & 34.1                  & 30.0                     & 47.5                  & 31.3                  \\ \hline
\multicolumn{1}{c|}{FitNet~\cite{fitnets}}                      & \multirow{3}{*}{\begin{tabular}[c]{@{}c@{}}R50\\ Faster R-CNN\\ (40.3)~\cite{faster_rcnn}\end{tabular}} & 34.2                     & 52.2                  & 36.1                  & 32.0                     & 49.3                  & 33.7                  \\
\multicolumn{1}{c|}{FGD~\cite{fgd}}                             &                                                                                                         & 35.4                     & 53.8                  & 37.3                  & 32.9                     & 50.5                  & 34.6                  \\
\multicolumn{1}{c|}{\textbf{HEAD\textsuperscript{\dag} (ours)}} &                                                                                                         & \textbf{36.0}            & \textbf{54.9}         & \textbf{38.4}         & \textbf{33.5}            & \textbf{51.6}         & \textbf{35.2}         \\ \hline
\multicolumn{1}{c|}{FitNet~\cite{fitnets}}                      & \multirow{3}{*}{\begin{tabular}[c]{@{}c@{}}R50\\ RepPoints\\ (38.6)~\cite{reppoints}\end{tabular}}      & 32.7                     & 50.9                  & 34.4                  & 30.3                     & 47.6                  & 31.7                  \\
\multicolumn{1}{c|}{FGD~\cite{fgd}}                             &                                                                                                         & 33.8                     & 52.1                  & 35.7                  & 31.2                     & 48.8                  & 32.5                  \\
\multicolumn{1}{c|}{\textbf{HEAD\textsuperscript{\dag} (ours)}} &                                                                                                         & \textbf{35.0}            & \textbf{53.8}         & \textbf{36.8}         & \textbf{32.5}            & \textbf{50.4}         & \textbf{34.3}         \\ \hline
\end{tabular}
\end{center}
\end{table}

\noindent\textbf{Comparison with homo-KD methods.}
In this section, HEAD is compared with previous homogeneous detection KD methods. 
Since these methods are originally proposed for homogeneous detectors, only the backbone mimicking part can be applied. 
For fairness, we do not use the CKD loss $\mathcal{L}^C$, but only use the AKD loss $\mathcal{L}^A$. 
We conduct experiments with two student architectures (RetinaNet~\cite{retinanet} and FCOS~\cite{fcos}), two student backbones (R18~\cite{resnet} and MNV2~\cite{mobilenetv2}), and two teacher choices (R50 Faster R-CNN~\cite{faster_rcnn} and R50 RepPoints~\cite{reppoints}). 
On all eight teacher-student pairs, \cref{tab:homo_sota} shows that HEAD outperforms the previous methods by a large margin. 
Specifically, under the guidance of R50 Faster R-CNN, our HEAD framework boosts the RetinaNet~\cite{retinanet} performance by $3.8$ and $4.0$ mAP for R18 and MNV2 backbones respectively. 
For FCOS, the performance gain is also prominent. 
We observe a $3.5$ mAP gain on both backbones when using R50 Faster R-CNN as the teacher. 
Interestingly, we observe that homo-KD methods degrade the student's performance in some cases when applied to heterogeneous detector pairs. 
Because the semantic gap between heterogeneous detectors is much larger than homogeneous detectors, the homo-KD methods are prone to over regularize the student, which causes this phenomena~\cite{pkth}.

\vspace{1mm}\noindent\textbf{Comparison with hetero-KD methods.}
We further compare our HEAD with the previous hetero-KD methods. 
When using Faster R-CNN as the teacher, we compare our HEAD with G-DetKD~\cite{gdetkd}. 
For RepPoints~\cite{reppoints} teacher, since the contrastive loss in G-DetKD is not applicable to such one-stage detector, we use its backbone mimicking loss (SGFI loss) only. 
For fairness, we disable our CKD loss as well. 
As shown in \cref{tab:hetero_sota}, our HEAD surpasses G-DetKD and SGFI on various teacher-student pairs.
Notice that Faster R-CNN uses the $2-6$ levels of the FPN~\cite{fpn} features, while RetinaNet, FCOS, and RepPoints use the $3-7$ levels.
The result suggests that the semantic-guided feature level matching mechanism cannot effectively bridge the semantic gap between the teacher and the student if both detectors use the same FPN levels. 
In contrast, our HEAD bridges the semantic gap by introducing assistants to homogenize the teacher-student pair, which results in over $2.5$ mAP gain on all scenarios.

\begin{table}[!t]
\caption{Comparison with heterogeneous KD methods. \textbf{\dag~indicates that only AKD losses are used in HEAD.}}
\label{tab:hetero_sota}
\begin{center}
\begin{tabular}{cc|ccc|ccc}
\hline
\multicolumn{2}{c|}{Student Backbone}                                                                                                                                       & \multicolumn{3}{c|}{R18~\cite{resnet}}                             & \multicolumn{3}{c}{MNV2~\cite{mobilenetv2}}                         \\ \hline
\multicolumn{1}{c|}{Method}                                     & Teacher                                                                                                   & mAP             & mAP\textsubscript{50} & mAP\textsubscript{75} & mAP             & mAP\textsubscript{50} & mAP\textsubscript{75} \\ \hline\hline
\multicolumn{1}{c|}{RetinaNet~\cite{retinanet}}                 & -                                                                                                         & 31.7          & 49.5                  & 33.5                  & 28.5          & 44.8                  & 29.9                  \\ \hline
\multicolumn{1}{c|}{G-DetKD~\cite{gdetkd}}                      & \multirow{2}{*}{\begin{tabular}[c]{@{}c@{}}R50 Faster R-CNN\\ (40.3)~\cite{faster_rcnn}\end{tabular}}   & 35.4          & 54.2                  & 37.9                  & 32.6          & \textbf{50.9}         & \textbf{34.5}         \\
\multicolumn{1}{c|}{\textbf{HEAD (ours)}}                       &                                                                                                           & \textbf{36.2} & \textbf{55.2}         & \textbf{38.8}         & \textbf{32.8} & 50.8                  & 34.4                  \\ \hline
\multicolumn{1}{c|}{SGFI~\cite{gdetkd}}                         & \multirow{2}{*}{\begin{tabular}[c]{@{}c@{}}R50 RepPoints\\ (38.6)~\cite{reppoints}\end{tabular}}        & 31.6          & 49.5                  & 33.2                  & 27.9          & 43.9                  & 29.2                  \\
\multicolumn{1}{c|}{\textbf{HEAD\textsuperscript{\dag} (ours)}} &                                                                                                           & \textbf{34.2} & \textbf{52.4}         & \textbf{36.6}         & \textbf{30.5} & \textbf{47.1}         & \textbf{32.3}         \\ \hline\hline
\multicolumn{1}{c|}{FCOS~\cite{fcos}}                           & -                                                                                                         & 32.5          & 50.9                  & 34.1                  & 30.0          & 47.5                  & 31.3                  \\ \hline
\multicolumn{1}{c|}{G-DetKD~\cite{gdetkd}}                      & \multirow{2}{*}{\begin{tabular}[c]{@{}c@{}}R50 Faster R-CNN\\ (40.3)~\cite{faster_rcnn}\end{tabular}} & 34.1          & 52.6                  & 36.3                  & 32.1          & 50.4                  & 33.6                  \\
\multicolumn{1}{c|}{\textbf{HEAD (ours)}}                       &                                                                                                           & \textbf{36.0} & \textbf{54.9}         & \textbf{38.4}         & \textbf{33.5} & \textbf{51.6}         & \textbf{35.2}         \\ \hline
\multicolumn{1}{c|}{SGFI~\cite{gdetkd}}                         & \multirow{2}{*}{\begin{tabular}[c]{@{}c@{}}R50 RepPoints\\ (38.6)~\cite{reppoints}\end{tabular}}       & 32.6          & 50.9                  & 34.4                  & 30.2          & 47.6                  & 31.7                  \\
\multicolumn{1}{c|}{\textbf{HEAD\textsuperscript{\dag} (ours)}} &                                                                                                           & \textbf{35.0} & \textbf{53.8}         & \textbf{36.8}         & \textbf{32.5} & \textbf{50.4}         & \textbf{34.3}         \\ \hline\hline
\multicolumn{1}{c|}{Faster R-CNN~\cite{faster_rcnn}}            & -                                                                                                         & 33.9          & 54.1                  & 36.3                  & 28.3          & 47.0                  & 29.5                  \\ \hline
\multicolumn{1}{c|}{G-DetKD~\cite{gdetkd}}                      & \multirow{2}{*}{\begin{tabular}[c]{@{}c@{}}R50 Cascade R-CNN\\ (43.5)~\cite{cascade_rcnn}\end{tabular}} & 36.1          & 57.3                  & 39.0                  & 33.4          & 54.2                  & 35.3                  \\
\multicolumn{1}{c|}{\textbf{HEAD (ours)}}                       &                                                                                                           & \textbf{36.7} & \textbf{58.0}         & \textbf{39.3}         & \textbf{33.8} & \textbf{54.4}         & \textbf{35.8}         \\ \hline
\end{tabular}
\end{center}
\end{table}

\vspace{1mm}\noindent\textbf{Comparison with teacher-free methods.}
Some methods have implemented teacher-free KD on object detection, such as MimicDet~\cite{mimicdet} and LabelEnc~\cite{labelenc}. 
To verify the superiority of our TF-HEAD among teacher-free methods without changing the structure of the student model, we only choose LabelEnc for comparison.
LabelEnc adopts a two-step training process, where both steps take $12$ epochs (1x) for training. Therefore, we use a 2x (24 epoch) training schedule for TF-HEAD. 
Three detection architectures with R50 backbones are explored, \ie RetinaNet~\cite{retinanet}, FCOS~\cite{fcos}, and Faster R-CNN~\cite{faster_rcnn}.
Our TF-HEAD uses two assistants for each architecture.
For RetinaNet, TF-HEAD uses FCOS$\star$ head and DH R-CNN head~\cite{double_head}.
For FCOS, RetinaNet head and DH R-CNN head are used.
For Faster R-CNN, we use RetinaNet head and FCOS$\star$ head to guide the RPN module.
As shown in \cref{tab:tf_sota}, HEAD improves baselines by $1.5$, $1.3$, and $1.1$ mAP, which is $67$\%, $63$\%, and $350$\% higher than the improvement of LabelEnc.

\begin{table}[!t]
\caption{Comparison with the teacher-free detection KD method. For fairness, we use R50~\cite{resnet} backbone and 2x (24 epoch) training schedule for all experiments. $\star$ indicates FCOS with improvements including center-sampling, normalization on bbox, centerness on regression branch, and GIoU.}
\label{tab:tf_sota}
\begin{center}
\begin{tabular}{c|c|c|c}
\hline
Method                   & RetinaNet~\cite{retinanet} & FCOS$\star$~\cite{fcos} & Faster R-CNN~\cite{faster_rcnn} \\ \hline
Baseline                 & 38.7                       & 41.0                    & 39.4                            \\
LabelEnc~\cite{labelenc} & 39.6                       & 41.8                    & 39.6                            \\
\textbf{TF-HEAD (ours)}  & \textbf{40.2}              & \textbf{42.3}           & \textbf{40.5}                   \\ \hline
\end{tabular}
\end{center}
\end{table}

\subsection{Ablation Study}

\vspace{1mm}\noindent\textbf{Effectiveness of HEAD.}
To evaluate the effectiveness of the assistant as a bridge between the teacher and the student, we conduct experiments without CKD loss. 
As show in \cref{tab:head}, HEAD\textsuperscript{\dag} improves R18 RetinaNet by $3.8$ mAP and MNV2 RetinaNet by $4.0$ mAP.
Adding CKD loss further brings $0.7$ mAP gain to R18 RetinaNet and $0.3$ mAP gain to MNV2 RetinaNet. 

\begin{table}[!t]
\caption{Ablation study of the HEAD. We use R50 Faster R-CNN as teacher and RetinaNet as student. \dag~indicates that only AKD losses are used in HEAD.}
\label{tab:head}
\begin{center}
\begin{tabular}{cc|ccc|ccc}
\hline
\multicolumn{2}{c|}{Student Backbone}                                                                                                                    & \multicolumn{3}{c|}{R18~\cite{resnet}}                           & \multicolumn{3}{c}{MNV2~\cite{mobilenetv2}}                       \\ \hline
\multicolumn{1}{c|}{Method}                     & Teacher                                                                                                & mAP           & mAP\textsubscript{50} & mAP\textsubscript{75} & mAP           & mAP\textsubscript{50} & mAP\textsubscript{75} \\ \hline
\multicolumn{1}{c|}{RetinaNet~\cite{retinanet}} & -                                                                                                      & 31.7          & 49.5                    & 33.5                    & 28.5          & 44.8                    & 29.9                    \\ \hline
\multicolumn{1}{c|}{\textbf{HEAD\textsuperscript{\dag}}} & \multirow{2}{*}{\begin{tabular}[c]{@{}c@{}}R50 Faster R-CNN\\ (40.3)~\cite{faster_rcnn}\end{tabular}} & 35.5          & 54.5                    & 37.9                    & 32.5          & 50.4                    & \textbf{34.4}           \\
\multicolumn{1}{c|}{\textbf{HEAD}}              &                                                                                                        & \textbf{36.2} & \textbf{55.2}           & \textbf{38.8}           & \textbf{32.8} & \textbf{50.8}           & \textbf{34.4}           \\ \hline
\multicolumn{1}{c|}{\textbf{HEAD\textsuperscript{\dag}}} & \multirow{2}{*}{\begin{tabular}[c]{@{}c@{}}R50 RepPoints\\ (38.6)~\cite{reppoints}\end{tabular}} & 34.2          & 52.4                    & \textbf{36.6}                    & 30.5          & 47.1                    & \textbf{32.3}           \\
\multicolumn{1}{c|}{\textbf{HEAD}}              &                                                                                                        & \textbf{34.3} & \textbf{52.8}           & 36.4           & \textbf{30.6} & \textbf{47.7}           & \textbf{32.3}           \\ \hline
\end{tabular}
\end{center}
\end{table}

\vspace{1mm}\noindent\textbf{Effectiveness of TF-HEAD.}
Here, we investigate the effectiveness of each distillation component in \cref{eq:tf_head_loss}.
We use FCOS head and R-CNN head as assistants to guide an R18 RetinaNet.
As shown in \cref{tab:tf_head}, our proposed TF-HEAD achieves $2.2$ mAP improvement over the original RetinaNet, demonstrating the effectiveness of our TF-HEAD structure.
FCOS assistant brings point-based knowledge for the backbone, boosting the baseline by $1.2$ mAP.
Then, CKD loss between the FCOS assistant and the RetinaNet head adds a $0.1$ mAP gain.
In contrast, the R-CNN assistant on its own brings $1.7$ mAP gain.
The CKD loss further adds $0.3$ mAP.
Using both assistants simultaneously reaches $33.9$ mAP.

\begin{table}[!t]
\caption{Ablation study of TF-HEAD. The student is R18 RetinaNet~\cite{retinanet}. We use FCOS head and R-CNN head as two assistants to guide the student.}
\label{tab:tf_head}
\begin{center}
\begin{tabular}{cccc|cccccc}
\hline
$\mathcal{L}^A_{FCOS}$ & $\mathcal{L}^C_{FCOS}$ & $\mathcal{L}^A_{R-CNN}$ & $\mathcal{L}^C_{R-CNN}$ & mAP           & mAP\textsubscript{50} & mAP\textsubscript{75} & mAP\textsubscript{s} & mAP\textsubscript{m} & mAP\textsubscript{l} \\ \hline
\multicolumn{1}{l}{}   & \multicolumn{1}{l}{}   & \multicolumn{1}{l}{}    & \multicolumn{1}{l|}{}   & 31.7          & 49.5                    & 33.5                    & 16.8                 & 34.7                 & 42.1                 \\ \hline
\textbf{\checkmark}    &                        &                         &                         & 32.9          & 51.2                    & 34.8                    & 17.8                 & 36.0                 & 43.5                 \\
\textbf{\checkmark}    & \textbf{\checkmark}    &                         &                         & 33.0          & 51.4                    & 35.0                    & 17.3                 & 36.3                 & 43.9                 \\ \hline
                       &                        & \textbf{\checkmark}     &                         & 33.4          & 52.1                    & 35.1                    & 18.1                 & 36.4                 & 44.8                 \\
                       &                        & \textbf{\checkmark}     & \textbf{\checkmark}     & 33.7          & 52.3                    & 35.6                    & 17.8                 & 36.9                 & 44.9                 \\ \hline
\textbf{\checkmark}    &                        & \textbf{\checkmark}     &                         & 33.8          & 52.5                    & 35.6                    & 18.2                 & 36.8                 & \textbf{45.2}                 \\
\textbf{\checkmark}    & \textbf{\checkmark}    & \textbf{\checkmark}     & \textbf{\checkmark}     & \textbf{33.9} & \textbf{52.7}           & \textbf{35.9}           & \textbf{18.5}        & \textbf{37.4}        & \textbf{45.2}        \\ \hline
\end{tabular}
\end{center}
\end{table}

\begin{figure}[!t]
  \begin{minipage}{0.45\linewidth} 
    \includegraphics[width=\linewidth]{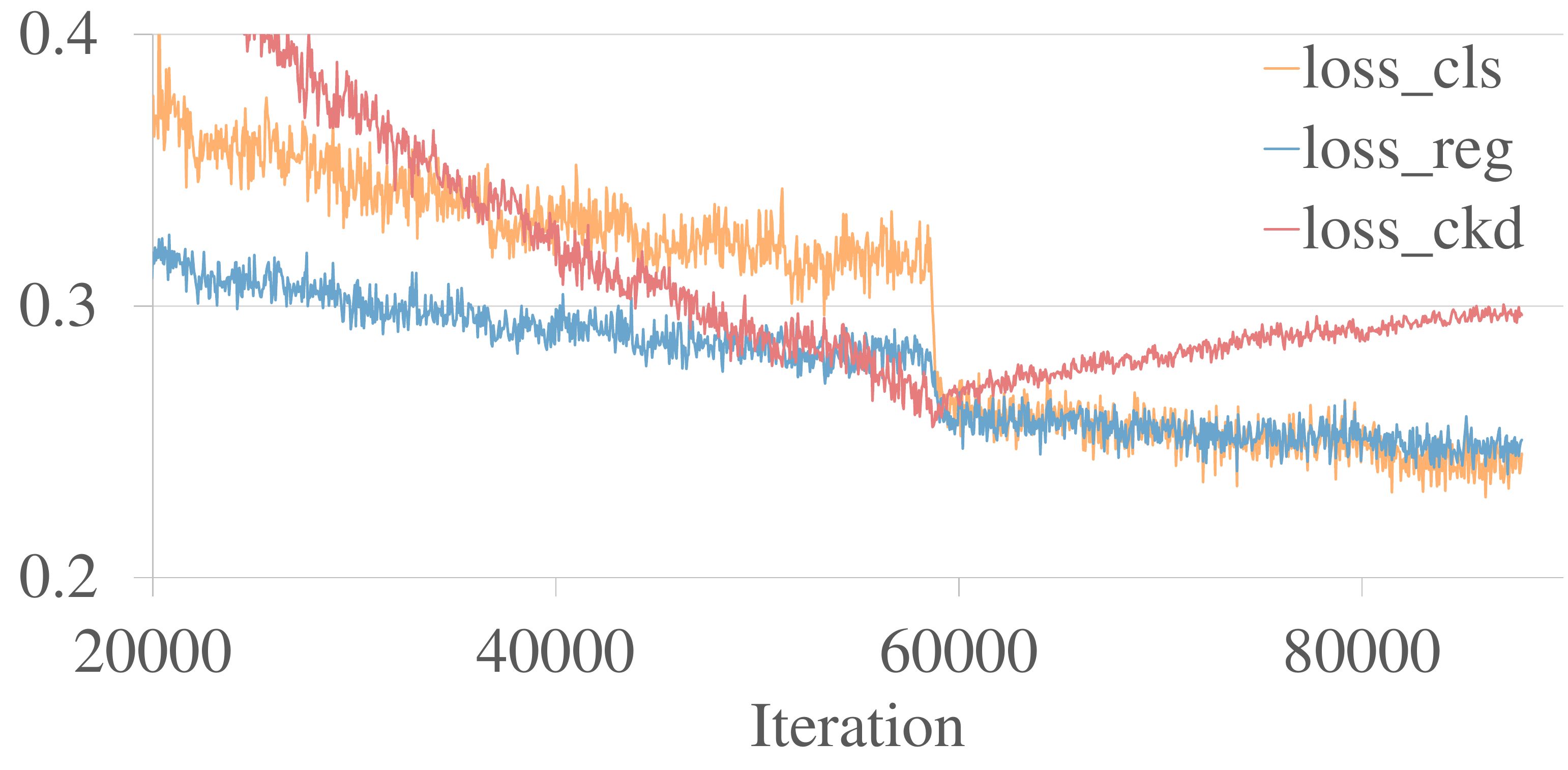} 
    \caption{Visualization of the CKD loss and ground truth losses.} 
    \label{fig:early_stop} 
  \end{minipage}
  \hfill
  \begin{minipage}{0.5\linewidth}
\captionof{table}{Early stopping the CKD losses increases the student performance by $0.1$ mAP. Repeated experiments confirm that it is not caused by random factors.}
\label{tab:early_stop}
\begin{center}
\begin{tabular}{c|cccc}
\hline
Early-stop          & mAP             & mAP\textsubscript{s} & mAP\textsubscript{m} & mAP\textsubscript{l} \\ \hline
                    & 33.8          & \textbf{18.5}      & 37.3               & 44.9               \\
\textbf{\checkmark} & \textbf{33.9} & \textbf{18.5}      & \textbf{37.4}      & \textbf{45.2}      \\ \hline
\end{tabular}
\end{center}
  \end{minipage}
\end{figure}

\vspace{1mm}\noindent\textbf{Early-stop to prevent misguidance from assistants.}
As shown in \cref{fig:early_stop}, from the $8$\textsuperscript{th} epoch, the CKD loss preternaturally increases while all other losses suddenly dropdown.
This phenomenon suggests that the CKD loss has stopped helping the student head since then.
Note that the $8$\textsuperscript{th} epoch is the first time to drop the learning rate.
Annealing KD~\cite{annealing_kd} believes that the detrimental effect is because the teacher disturbs the student from learning the ground truth labels.
After the learning rate drops at the $8$\textsuperscript{th} epoch, the optimization direction of the assistant and the student head start to diverge, which enlarges the semantic gap in between.
From this moment, CKD loss starts to over regularize~\cite{pkth} the student.
Intuitively, the student head and the assistant have different architectures, so they should predict objects differently.
Forcing the student head to mimic the assistants will impede it from learning knowledge by itself effectively.
Therefore, we use early-stop to ensure that the student and assistants converge to their local optima.
\Cref{tab:early_stop} shows that early-stop slightly improves the performance. 

\subsection{Visualization}

\begin{figure}[!t]
  \centering
  \includegraphics[width=0.95\linewidth]{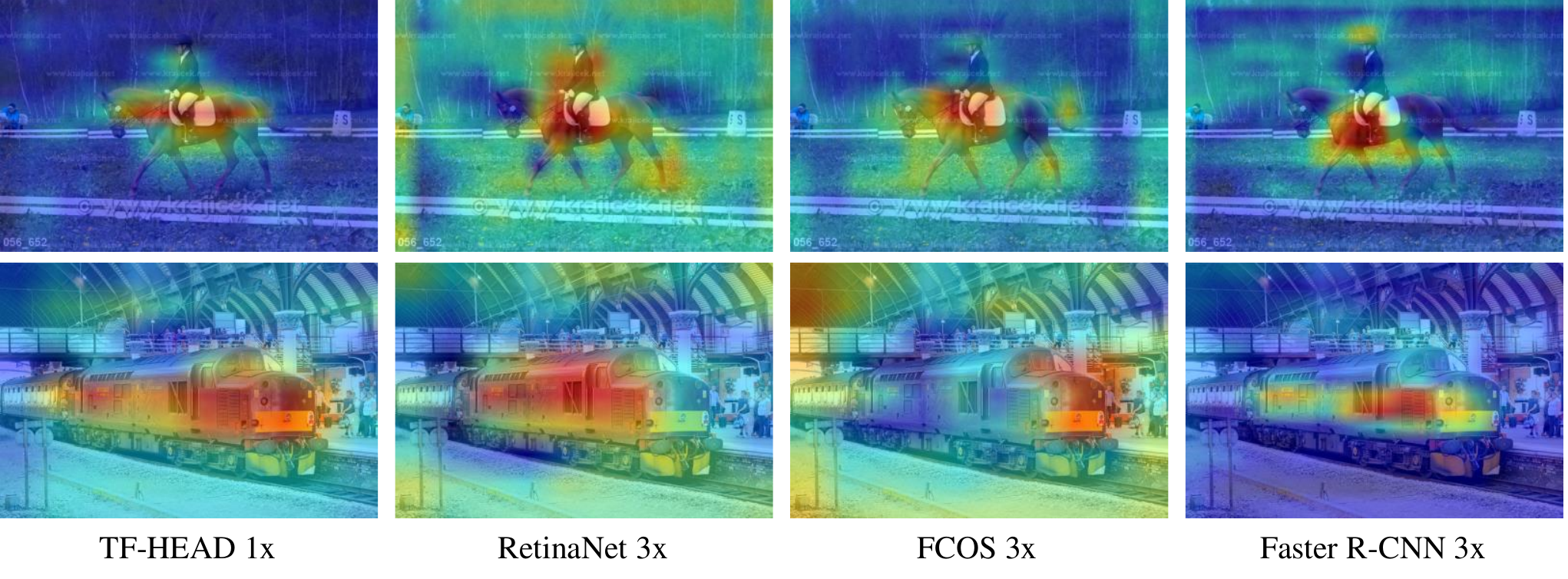}
  \caption{
    Visualization of backbone features from TF-HEAD, RetinaNet~\cite{retinanet}, FCOS~\cite{fcos}, and Faster R-CNN~\cite{faster_rcnn}.
    TF-HEAD highlights more foreground area with higher intensity, while the background remains inactivated.
  }
  \label{fig:vis}
\end{figure}

\vspace{1mm}\noindent\textbf{Visualization of backbone feature maps.}
By visualizing the feature maps, We verify that TF-HEAD trains stronger backbones. 
As shown in \cref{fig:vis}, feature maps generated by HEAD accurately identify the regions containing objects, showing that knowledge of RetinaNet~\cite{retinanet}, FCOS~\cite{fcos}, and Faster R-CNN~\cite{faster_rcnn} can complement each other. 
Therefore, the backbone of TF-HEAD generates more informative feature maps, which further helps the student be more accurate.

\begin{figure}[t]
  \centering
  \includegraphics[width=0.8\linewidth]{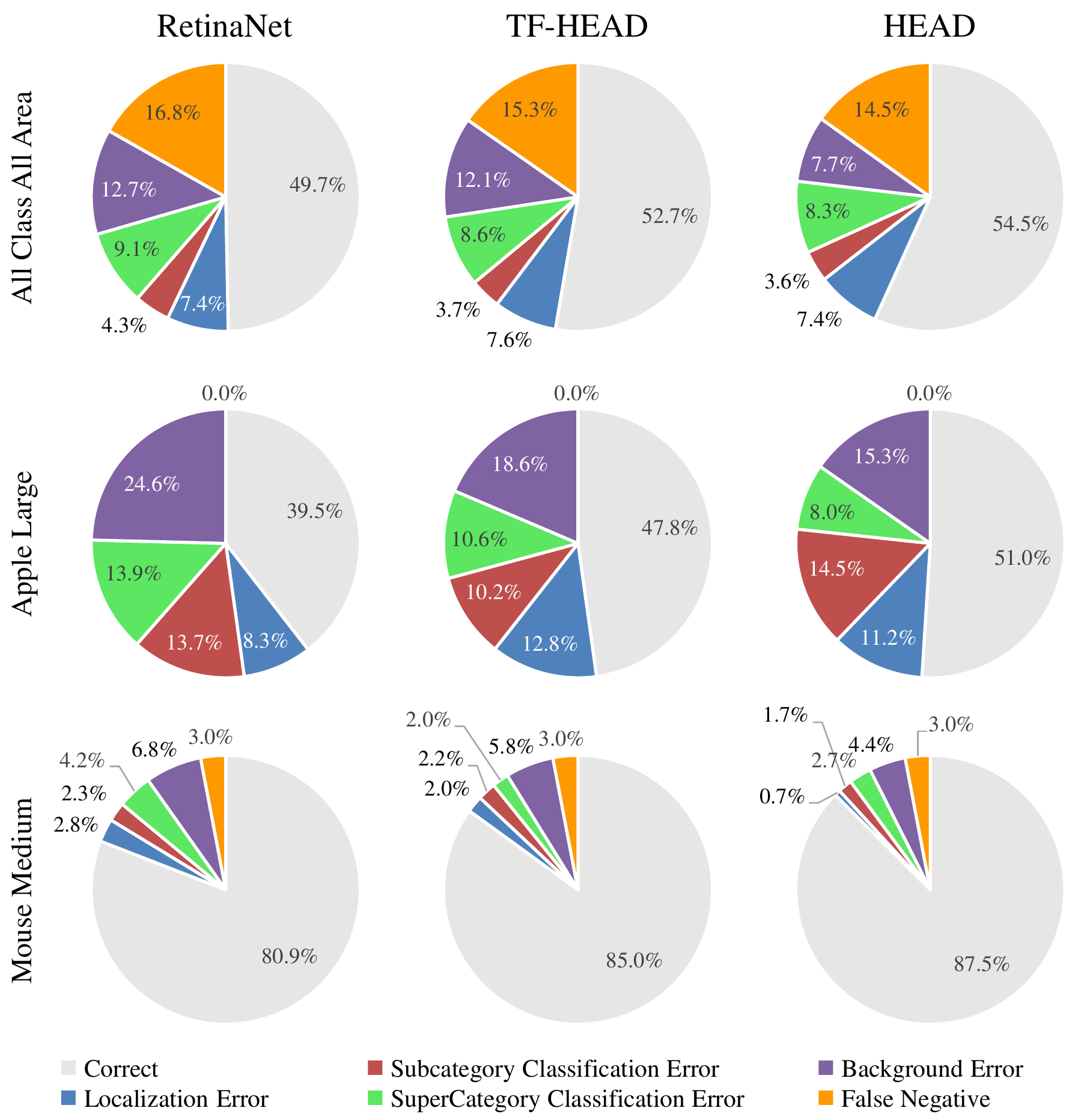}
  \caption{COCO error analysis using tool from \cite{coco_error_analysis}.}
  \label{fig:error}
\end{figure}

\vspace{1mm}\noindent\textbf{Visualization of COCO error analysis.} 
\Cref{fig:error} presents analysis on \textit{All Class} and two randomly selected classes.
From RetinaNet to TF-HEAD, the \textit{Background} error and the \textit{False Negative} error decreases prominently, suggesting that the student has learned knowledge from heterogeneous detectors to make more accurate predictions.
As a result, the \textit{Correct} rate increases prominently.
Using the pretrained teacher detector, HEAD further pushes the accuracy higher, with the \textit{Background} error and the \textit{False Negative} error even lower.

\section{Conclusion}

In this paper, we investigate KD among heterogeneous object detectors and find that the semantic gap between heterogeneous models is responsible for the difficulty of hetero-KD. Based on the observation, we design a simple yet effective HEtero-Assists Distillation (HEAD) mechanism. HEAD bridges the semantic gap between heterogeneous detectors via an adaptive assistant, thus simplifying to a homo-KD problem. For situations when the pretrained teachers are not available, we further propose a teacher-free method named TF-HEAD. Extensive experiments demonstrate the effectiveness of our framework.

\vspace{1mm}\noindent\textbf{Acknowledgement.} 
This work was partly supported by the National Natural Science Foundation of China (62122010, 61876177), the Fundamental Research Funds for the Central Universities, and the Key Research and Development Program of Zhejiang Province (2022C01082). 

\bibliographystyle{splncs04}
\bibliography{library}
\end{document}